\documentclass[11pt,a4paper]{article}
\usepackage[hyperref]{acl2021}
\usepackage{times}
\usepackage{latexsym}

\usepackage{microtype}

\usepackage{latexsym}
\usepackage{array}
\usepackage{color}
\usepackage{bm}
\usepackage{enumitem}
\usepackage{booktabs}
\usepackage{amssymb}
\usepackage{amsfonts}
\usepackage{amsmath}
\usepackage{multirow}
\usepackage{graphicx}
\usepackage{amsfonts}
\usepackage[linesnumbered,ruled,vlined]{algorithm2e}
\usepackage{colortbl}
\usepackage{CJKutf8}
\usepackage{url}
\usepackage{textcomp}
\usepackage[normalem]{ulem}

\usepackage{CJKutf8}
\usepackage[utf8]{inputenc}
\usepackage[T1]{fontenc}
\newcommand{\ja}[1]{\begin{CJK}{UTF8}{ipxm}#1\end{CJK}}

\newcommand{\todo}[1]{}
\renewcommand{\todo}[1]{{\color{red} TODO: {#1}}}
\newcommand{\chousa}[1]{}
\renewcommand{\chousa}[1]{{\color{magenta} Chousa: {#1}}}
\newcommand{\morishita}[1]{}
\renewcommand{\morishita}[1]{{\color{purple} Morishita: {#1}}}

\newcommand{\entoja}{En\ensuremath{\rightarrow}Ja}
\newcommand{\jatoen}{Ja\ensuremath{\rightarrow}En}

\newcommand{\sep}{$ \langle \texttt{sep} \rangle $}
\newcommand{\eos}{$ \langle \texttt{eos} \rangle $}

\newcommand{\baseline}{\textsc{Base}}

\aclfinalcopy 

\title{Input Augmentation Improves Constrained Beam Search \\for Neural Machine Translation: NTT at WAT 2021}

\author{
    Katsuki Chousa\thanks{\ \ Equal contribution.} \and
    Makoto Morishita\footnotemark[1] \\
    NTT Communication Science Laboratories, NTT Corporation \\
    2-4, Hikaridai, Seika-cho, Soraku-gun, Kyoto, 619-0237, Japan \\
    \texttt{\{katsuki.chousa.bg, makoto.morishita.gr\}@hco.ntt.co.jp} \\
}

\date{}

\begin{document}
\maketitle
\begin{abstract}
This paper describes our systems that were submitted to the restricted translation task at WAT 2021.
In this task, the systems are required to output translated sentences that contain all given word constraints.
Our system combined input augmentation and constrained beam search algorithms.
Through experiments, we found that this combination significantly improves translation accuracy and can save inference time while containing all the constraints in the output.
For both \entoja\ and \jatoen, our systems obtained the best translation performances in both automatic and human evaluations.
\end{abstract}

\section{Introduction}
\begin{figure}[t]
    \centering
    \includegraphics[width=\linewidth]{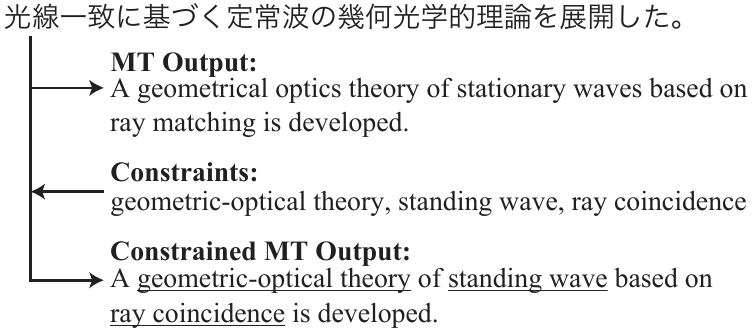}
    \caption{Overview of the restricted translation task}
    \label{fig:task_overview}
\end{figure}

This year, we participated in the restricted translation task at WAT 2021 \citep{nakazawa-etal-2021-overview}, in which we were asked to control a model so that the translation output would contain specified terms.
Although the recent neural machine translation (NMT) model achieves excellent performance, controlling its output is still a challenging task.
Figure~\ref{fig:task_overview} shows an overview of the task.
Each sentence includes the target words (constraints) that must be contained in the output.
We believe this task reflects a critical function, especially in practical applications.
For example, users may want to control the translation of technical terms or proper nouns.

Several works have tried to control the NMT outputs, and these works can be divided into two categories: {\it hard} and {\it soft} methods.
The hard lexically constrained method guarantees that all the target words are in the output.
Current works achieve this by modifying the beam search algorithm to find the hypothesis that contains all of the target words~\cite{hokamp-liu-2017-lexically,post-vilar-2018-fast}.
The hard method guarantees all constraints are satisfied, but its translation performance is sometimes lower than the conventional NMT.
This is because it requires all given target words to be contained in the decoding step, which may disrupt the model inference.

The soft lexically constrained method, on the other hand, does not guarantee that all target words are contained in the output.
These methods usually modify or augment the input of the NMT model and try to output the given target words without changing the decoding algorithm~\cite{song-etal-2019-code,chen2020lexical_leca}.
Its decoding speed is usually faster than the hard method, but some of the constraints may not be satisfied.

Our submission aims to contain all of the specified target words with high translation accuracy.
To achieve this goal, we applied both input augmentation and constrained beam search algorithms.
To the best of our knowledge, this is the first work that combines these two methods.
Through experiments, we found that this combination achieves quite high translation performance while containing all target words in the output and saving inference time.
We submitted the systems to the English-to-Japanese (\entoja) and Japanese-to-English (\jatoen) tasks, and we were ranked first in both language pairs in terms of BLEU scores and human evaluations.

\section{Task Definition}\label{sec:task_definition}
Suppose we have a source sentence $X=(x_1, x_2, \ldots, x_S)$ with $S$ tokens and a target sentence $Y=(y_1, y_2, \ldots, y_T)$ with $T$ tokens.
In a conventional machine translation approach, the problem of translation from $X$ to $Y$ can be solved by finding the best target sentence that maximizes the conditional probability
\begin{equation}
    p(Y \mid X) = \prod_{t=1}^{T}{p(y_t \mid y_{<t}, X)}.
\end{equation}

In the restricted translation task, lists of target words are provided to represent word restrictions, and systems are required to output translations that contain all of the target words in each list.
Here, the problem of translation with word constraints can be defined as
\begin{equation}
    p(Y \mid X, C) = \prod_{t=1}^{T}{p(y_t \mid y_{<t}, X, C)},
\end{equation}
where $C = (C_1, C_2, \ldots, C_N)$ is the provided word constraints with $N$ phrases, and the constraints are given in random order.

The performance of systems in this task is evaluated through two metrics:
\begin{itemize}
    \item Translation accuracy: BLEU \citep{papineni-EtAl:2002:ACL} is used for evaluation in this task.
    \item Consistency score: The percentage of sentences that correctly contain the given constraints over the entire test set.
\end{itemize}
For the final ranking, the combined score of the above metrics is calculated as follows:
\begin{enumerate}
    \item If the translation does not contain all of the constraints based on exact matching, replace the translation with an empty string.
    \item Calculate BLEU scores with modified translations.
\end{enumerate}

\section{Data}
\subsection{Provided Data}
In this task, we were asked to translate an English/Japanese scientific paper.
As the in-domain training data, organizers provided ASPEC~\cite{nakazawa16aspec}, which contains three million parallel sentences.
Since this corpus is ordered by the sentence-alignment quality, the sentences at the end might be noisy.
Following a previous work~\cite{morishita17wat}, we used only the first two million sentences as parallel sentences.
We treated the final one million sentences as monolingual data and created a synthetic corpus~\cite{sennrich16acl}.
Based on a previous analysis~\cite{morishita:2019:ntt}, we forward-translated it for the Japanese-English task and back-translated it for the English-Japanese task.

\subsection{Other Resources}
We also trained the model with additional resources.
As an additional parallel corpus, we used JParaCrawl~\cite{morishita20lrec}, which contains 10 million sentence pairs.

We also used CommonCrawl provided by the WMT 2020 news shared task~\cite{barrault20wmt} as additional monolingual data.
For CommonCrawl data, we chose the ten million English and Japanese sentences that are similar to the scientific domain based on the language model trained with ASPEC~\cite{moore-lewis10acl}.
Then we further filtered out the following noisy sentences: (1) non-English/Japanese sentences with CLD2~\footnote{\url{https://github.com/CLD2Owners/cld2}}, (2) excessively long sentences (more than 250 subwords), (3) sentences that contain out-of-vocabulary characters.
After cleaning, we kept 7.9 million English and 9.2 million Japanese sentences.
We then back-translated these sentences with the NMT model trained with ASPEC to make a synthetic corpus.

\section{System Details}
\subsection{Base Model and Hyperparameters} 
\label{sec:base_model}
\begin{table}[t]
  \centering
  \small
  \tabcolsep=2pt
  \begin{tabular}{lp{43mm}}
  \toprule
    Architecture            &  Transformer (big) \\
    Tied-embeddings         &  Tied the encoder/decoder embeddings and the decoder output layer\\
     Optimizer              &   Adam ($\beta_{1}=0.9, \beta_{2}=0.98, \epsilon=1\times10^{-8}$)~\cite{kingma14adam}  \\
     Learning Rate Schedule &   Inverse square root decay     \\
     Warmup Steps           &   4,000  \\
     Max Learning Rate      &   0.001  \\
     Dropout                &   0.3  \\
     Gradient Clipping      &   1.0  \\
     Label Smoothing        &   $\epsilon_{ls}=0.1$~\citep{szegedy:2016:rethinking}     \\
     Mini-batch Size        &   512,000 tokens~\cite{ott18scaling}\\
     Number of Updates      &   8,000 steps \\
     Averaging              &   Save checkpoint for every 100 steps and take an average of last 8 checkpoints \\
  \bottomrule
  \end{tabular}
  \caption{List of hyperparameters}
  \label{tab:hyper-parameter}
\end{table}

As a baseline system, we employed the Transformer model with the big setting~\cite{vaswani17transformer}.
Table~\ref{tab:hyper-parameter} shows the detailed settings and hyperparameters.
As an NMT implementation, we used {\tt fairseq}~\cite{ott19fairseq}, and modified it in the following experiments.

\subsection{Lexically Constrained Decoding}\label{sec:lcd}
We used the lexically constrained decoding (LCD) technique \citep{hokamp-liu-2017-lexically, post-vilar-2018-fast} to incorporate constraints at decoding time.
In this task, the translations that do not satisfy the constraints lead to a substantial decrease in the final score.
This technique is a hard lexically constrained method that uses grid beam search algorithm, and it guarantees that all word constraints appear in the target sentence.

\begin{table}[tbp]
    \centering
    \tabcolsep=2pt
    \begin{tabular}{lrrr}\toprule
        {\bf Setting} & {\bf BLEU} & {\bf Term\%} & {\bf Sent}\% \\ \midrule
        \baseline & 29.4 & 50.80 & 23.3 \\
        \ + LCD (beam=60) & 24.0 & 94.40 & 85.3 \\
        LeCA & 42.2 & 87.64 & 72.02 \\
        \ + LCD (beam=30) & 43.9 & 94.34 & 85.21 \\\bottomrule
    \end{tabular}
    \caption{Comparison of translation accuracy and consistency score for each setting on \jatoen.}
    \label{tab:term_sent_per}
\end{table}
To evaluate the effectiveness of this technique, we compared the baseline model (\baseline) and the baseline with LCD (\baseline+LCD).
Here, we used two metrics for the consistency score: term\% is the percentage of constraints that are correctly generated in the translations, and sent\% is the percentage of sentences that contain all given constraints.
Table~\ref{tab:term_sent_per} shows that the \baseline+LCD significantly improves both term\% and sent\% on \jatoen.
The reason why the two consistency scores of \baseline+LCD are not 100\% is due to the normalization on the tokenization, and this can be addressed by post-processing (\S\ref{sec:post-processing}).

However, \baseline+LCD decreased the translation accuracy of the model.
In preliminary experiments with the baseline models, we also found that the beam size needs to be larger than 60 to successfully generate all the constraints in this task.
This is because the translations contain much repetition and the model never finishes generation before reaching the maximum output length.

\subsection{LExical-Constraint-Aware NMT} 
\label{sec:leca}
\begin{figure*}[t]
    \centering
    \includegraphics[width=\linewidth]{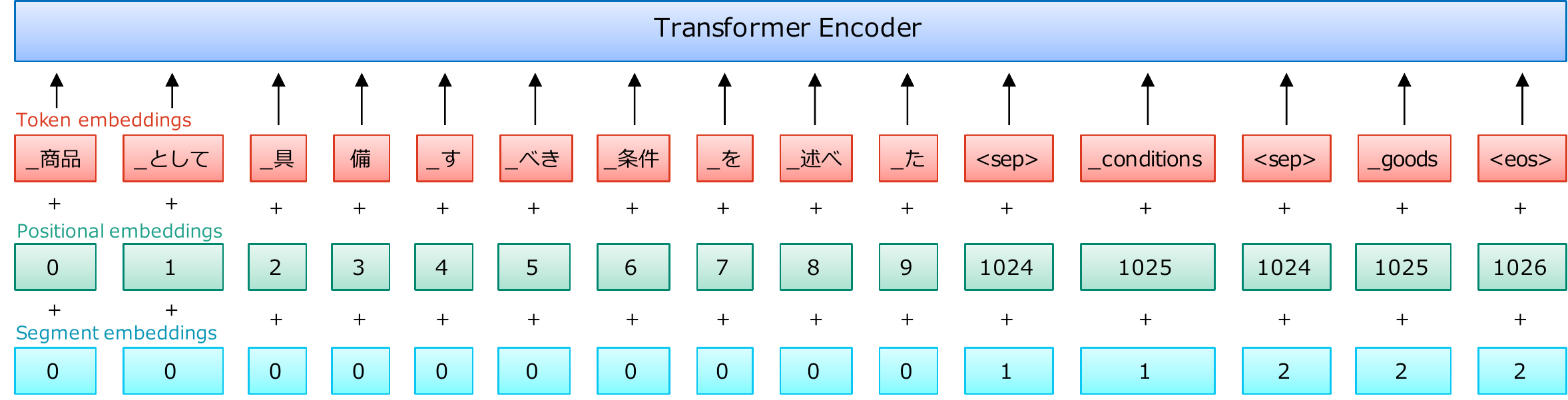}
    \caption{Input representation of the LeCA model.}
    \label{fig:leca_embed}
\end{figure*}
To ease the problem in LCD, we used the Lexical-Constraint-Aware NMT (LeCA) model~\citep{chen2020lexical_leca}, whose input is augmented by concatenating constraints and the source sentence together.
This method can inform the model of what constraints are given before decoding time, and thus the model can properly decide where to output a constraint.
LeCA is a one of the soft lexically constrained methods, which do not guarantee all constraints are in the output.
However, in combination with LCD, we can guarantee the model always satisfies the constraints while keeping or improving the translation performance.

The input is constructed by concatenating the source sentence $X$ and each phrase $C_i$ in the constraints $C$ with a separator symbol \sep, as follows:
\begin{equation}
    [ X, \langle \texttt{sep} \rangle, C_1, \langle \texttt{sep} \rangle, C_2, \ldots, C_N, \langle \texttt{eos} \rangle ], \label{eq:leca_input}
\end{equation}
where {\eos} is the symbol indicating the end of the sentence.

To construct the input at training time, \citet{chen2020lexical_leca} proposed a method that dynamically samples constraints from a reference sentence.
They first sampled the number of constrained words $k$, and then they randomly sampled $k$ target words (not subwords) as constraints from the reference.
Here, we sampled the number of constrained words $k$ from $0$ to $14$ following the distribution that is $p=0.4$ for $0$ and $p=0.6/14(=0.04)$ for the other ones.
The high probability for no constraint is to maintain the translation performance for unconstrained settings.

To handle such a source sequence, this method modifies the input representation of the encoder to distinguish the source sentence and each constraint.
This representation is composed of three types of learned embeddings: token embeddings, positional embeddings, and segment embeddings, as shown in Fig. \ref{fig:leca_embed}.
The position of each constraint starts from the maximum length of the source sentences to avoid overlapping with the sentence.
We assigned different values for the source sentence and each constraint and fed it to the model with the segment embeddings.
This method also introduces a pointer network architecture\cite{vinyals-et-at-2015-pointer,see-etal-2017-get} that helps to generate constraints by copying from the source sequence.
Finally, we updated the models with 10,000 steps for \jatoen\ and 12,000 steps for \entoja\ and set the beam size to 30 for LCD.

We evaluated the effectiveness of LeCA and LeCA with LCD (LeCA+LCD).
Table~\ref{tab:term_sent_per} shows that LeCA achieved high translation accuracy and consistency scores.
The input of both LeCA and \baseline+LCD are the same, but the translation accuracy of LeCA is significantly better than that of \baseline+LCD.
Moreover, LeCA+LCD with a small beam size improves the translation accuracy and satisfies all of the constraints.
This implies that inputting both a source sentence and constraints as source sequence is very effective for improving the performance in this task.

\subsection{Pre-process} 
\begin{table}[t]
    \centering
    \begin{tabular}{lrrr}\toprule
        \multicolumn{1}{l}{{\bf Tokenizer} }& {\bf BLEU} & {\bf Term\%} & \bf{Sent\%} \\ \midrule
        MeCab + ipadic & 44.8 & 68.67 & 43.87 \\
        MeCab + NEologd & 46.5 & 72.35 & 49.39 \\ \bottomrule
    \end{tabular}
    \caption{Comparison of translation performance when changing the dictionary of tokenizer on \entoja. The model setting is LeCA with a few updates.}
    \label{tab:mecab_dictionary}
\end{table}
Since constraints that are sampled from the reference are given as not a subword but a word, we need to separate the sentence into words.
To do this, we first tokenized both the input and output sentences.
For English, we simply applied the tokenizer scripts available in the \texttt{Moses} toolkit~\citep{koehn07moses}.
We used the Moses \texttt{truecaser} when the target language is English.
For Japanese, we use the \texttt{MeCab} tokenizer~\citep{kudo06mecab} with the \texttt{mecab-ipadic-NEologd}~\citep{sato2015mecabipadicneologd} dictionary.
This dictionary contains many neologisms and thus it helps in handling named entities or technical terms, which are included in ASPEC but cannot be tokenized correctly using the default system dictionary.
We compared the LeCA performance of \texttt{mecab-ipadic-NEologd} with the default system dictionary on an \entoja\ task.
Table~\ref{tab:mecab_dictionary} shows that \texttt{mecab-ipadic-NEologd} significantly improved translation accuracy and consistency scores.
We confirmed that using \texttt{mecab-ipadic-NEologd} is the best option for LeCA on this task.

Then, we trained subword encoding models using the \texttt{sentencepiece} implementation~\citep{kudo:2018:sentencepiece}.
According to an earlier work \citep{morishita:2019:ntt}, a smaller vocabulary size (e.g., 4,000) is empirically superior to the commonly used ones (e.g., 32,000).
On the other hand, larger vocabulary size is preferred for an LCD to keep the number of constraint tokens small.
This is because a large number of tokens requires a large beam size of the LCD and increases the inference time.
Finally, we found in a preliminary experiment that a vocabulary size of 32,000 achieved the best results, so we used a joint subword vocabulary with 32,000 tokens.
For training data, we applied the Moses \texttt{clean-corpus-n} scripts to remove sentence pairs that are either too long or too different int their lengths\footnote{We set the minimum length to 1, the maximum length to 250, and the maximum ratio of lengths to 9.}.

\subsection{Fine-Tuning and Data Selection} \label{sec:finetune_synth}
The synthetic corpora (e.g., ASPEC last 1M and CommonCrawl) contain noisy sentence pairs, and the domain of JParaCrawl is different from that of ASPEC, a scientific paper domain.
We used these corpora to make the translations more fluent.
The model was initially pre-trained with these corpora and the first 2M sentence pairs of ASPEC for 12,000 updates.
We then fine-tuned the pre-trained model using only the first 2M sentence pairs of ASPEC for 2,000 steps.
For the pre-training, we oversampled ASPEC three-times to keep roughly the same number of sentences as the synthetic corpora.

\begin{table}[t]
    \centering
    \begin{tabular}{lrr}\toprule
         & \multicolumn{2}{c}{\bf{BLEU}} \\
        \multicolumn{1}{l}{\bf{Setting}} & \bf{\jatoen} & \bf{\entoja} \\ \midrule
        ASPEC 2M & 44.34 & ---\footnotemark \\
        \ \ + synth 1M & 44.26 & 56.57 \\
        \ \ after pre-training & 44.28 & 56.47 \\ \bottomrule
    \end{tabular}
    \caption{Effectiveness of fine-tuning. The model settings are LeCA+LCD.}
    \label{tab:finetuning}
\end{table}
\footnotetext{
In a preliminary experiment on \entoja, we found that a model using synthetic data was superior to that using only ASPEC 2M.
However, we did not compare the three settings under the same conditions.
}
We searched for an effective setting to use the training data.
Table~\ref{tab:finetuning} shows the results.
The model using only ASPEC 2M for \entoja\ and the model using ASPEC 2M and forward-translated ASPEC last 1M for \jatoen\ achieved the highest translation accuracies.
For both \entoja\ and \jatoen, the models trained on ASPEC 2M after pre-training achieved comparable results to the best ones.
Since these models are trained on large amounts of parallel sentence pairs, they might be expected to produce more natural output than the best ones and thus be preferred by humans.
Therefore, we decided to submit these four models for human evaluation.

\subsection{Ensemble}\label{sec:ensemble}
We applied a model ensemble technique to improve the translation accuracy.
First, eight models were trained with different random seeds.
We then computed the average scores of these models and generated hypotheses based on these scores using beam search decoding.

\begin{table}[t]
    \centering
    \begin{tabular}{lrr}\toprule
         & \multicolumn{2}{c}{\bf{BLEU}} \\
        \bf{Model type} & \bf{\entoja} & \bf{\jatoen} \\ \midrule
        Single model & 55.49 & 43.44 \\
        8 Ensemble & 56.57 & 44.34 \\ \bottomrule
    \end{tabular}
    \caption{Effectiveness of ensembling models. The model settings are LeCA+LCD.}
    \label{tab:ensemble}
\end{table}
Table~\ref{tab:ensemble} shows the effectiveness of ensembling models.
Ensembling the eight models shows a significant improvement over the single model.

\subsection{Post-processing}
\label{sec:post-processing}
For the submission, we need to match the tokenization to the reference constraints.
To achieve this, we fixed the terms that are not matched to the constraints due to tokenization issues.
Specifically, for each unmatched constraint, we removed spaces in both the output and the constraint, and then replaced the constraint in the output with the reference-spaced constraint.
In some cases, we found that constraints may contain out-of-vocabulary (OOV) characters, resulting in translation failure\footnote{We found that two percent of the lines in the test set include OOV characters.}.
The model outputs the special OOV tokens for these sentence, and thus we replaced them with correct characters in the reference constraint.

\section{Official Results}
\begin{table*}[t]
    \centering
    \begin{tabular}{llrr}\toprule
         & & \multicolumn{2}{c}{\bf{BLEU}} \\
        {\bf ID} & \multicolumn{1}{l}{\bf{Setting}} & \multicolumn{1}{r}{\bf{\entoja}} & \multicolumn{1}{r}{\bf{\jatoen}} \\ \midrule
        (a) & \baseline\ (\S\ref{sec:base_model}) & 44.64 & 29.30 \\
        (b) & \baseline\ + LCD (\S\ref{sec:lcd}) & 45.38 & 23.22 \\
        (c) & LeCA (\S\ref{sec:leca}) & 53.79 & 41.88 \\
        (d) & LeCA + LCD & 55.49 & 43.33 \\
        (e) & (d) $\times$ 8 ensemble (\S\ref{sec:ensemble}) & \textbf{56.57} & \textbf{44.34} \\
        (f) & [(d) + fine-tuning (\S\ref{sec:finetune_synth})] $\times$ 8 & 56.47 & 44.28 \\
        \bottomrule
    \end{tabular}
    \caption{The performance of the submitted systems. According to \S\ref{sec:finetune_synth}, we used only ASPEC 2M for \entoja\ and ASPEC 2M + synth 2M for \jatoen. For \entoja, we show BLEU scores with \texttt{MeCab} tokenizer. Bold values indicate the highest score in each column.}
    \label{tab:official_bleu}
\end{table*}

Table~\ref{tab:official_bleu} shows the automatic evaluated performance of our systems on the test set.
These scores were measured in the evaluation server\footnote{\url{http://lotus.kuee.kyoto-u.ac.jp/WAT/evaluation/index.html}}.
The best systems improved the BLEU score by +11.93 pts for \entoja\ and +15.04 pts for \jatoen~against the \baseline.
Our systems achieved the best BLEU score for both \entoja\ and \jatoen\ subtasks.

\begin{table*}[t]
    \centering
    \begin{tabular}{ccccccc} \toprule
         & \multicolumn{2}{c}{{\bf Automatic Eval.}} & \multicolumn{4}{c}{{\bf Human Eval.}} \\\cmidrule(lr){2-3} \cmidrule(lr){4-7}
        {\bf Language pair} & {\bf Final score} & {\bf (Rank)} & {\bf DA} & {\bf (Rank)} & {\bf CA} & {\bf (Rank)} \\ \midrule
        \entoja & 57.2 & (1) & 77.5 & (1) & 79.7 & (1) \\
        \jatoen & 44.1 & (1) & 75.6 & (1) & 74.4 & (1) \\ \bottomrule
    \end{tabular}
    \caption{Official results of our team. The definition of the final score is described in \S\ref{sec:task_definition}. Human evaluations are based on source-based direct assessment (DA) \citep{cettolo17iwslt,federmann-2018-appraise} and source-based contrastive assessment (CA) \citep{sakaguchi-van-durme-2018-efficient,federmann-2018-appraise}.}
    \label{tab:official_results}
\end{table*}
Table~\ref{tab:official_results} shows the official results of our systems\footnote{The results of all participants are reported in \url{https://sites.google.com/view/restricted-translation-task/\#h.g3vfoh2oljpq}}.
For both \entoja\ and \jatoen, our systems achieved the best scores in the final ranking. 
Our submissions did not drop the scores from the BLEU, while the other participants dropped it. 
This means that our team only succeeded in implementing systems whose translation output could contain all the specified terms. 
Our systems also achieved the best performance in terms of human evaluations for both \entoja\ and \jatoen.
Notably, our scores are better than the reference ones even for \jatoen.
This implies that constrained translation can yield human-parity performance when the system can receive appropriate terms in the target language.

\section{Analysis}

\begin{figure*}[t]
\centering
\footnotesize
\begin{tabular}{ll}
\toprule
\textbf{Source} & \ja{分路巻線のみに補助巻線を持つ超電導単相単巻変圧器を試作した。} \\ \midrule
\multirow{2}{*}{\textbf{Reference}} & \uline{Superconductivity single phase auto‐transformer} with \uline{auxiliary winding} only at the \uline{shunt winding} was\\
& produced experimentally.\\ \midrule
\textbf{Constraints} & shunt winding, auxiliary winding, superconductivity single phase auto‐transformer\\ \midrule
\multirow{7}{*}{\textbf{Base+LCD}} & We have developed a \uwave{superconducting single ‐ phase transformer} with \uline{auxiliary winding}s only in the \uline{shunt}\\
                                   & \uline{winding}s, in which the \uline{auxiliary winding}s are connected to the \uline{shunt winding}s of \uwave{the single ‐ phase}\\
                                   & \uwave{transformer}, and the \uline{auxiliary winding}s are connected to the \uline{shunt winding}s of the \uwave{single ‐ phase transformer}\\
                                   & with \uline{auxiliary winding}s of the \uline{auxiliary winding}s of the \uline{auxiliary winding}s of the \uline{auxiliary winding}s of the\\
                                   & \uline{auxiliary winding}s of the \uline{auxiliary winding}s of the \uline{auxiliary winding}s of the \uline{auxiliary winding}s of the \\
                                   & \uline{auxiliary winding}s of the \uline{auxiliary winding}s. \uline{Superconductivity single phase auto ‐ transformer} is assisted\\
                                   & by the \uline{auxiliary winding}s of the \uline{auxiliary winding}s. \\ \midrule
\multirow{2}{*}{\textbf{LeCA+LCD}} & A \uline{superconductivity single phase auto‐transformer} with \uline{auxiliary winding} only in the \uline{shunt winding} was\\
& produced experimentally.\\
\bottomrule
\end{tabular}
\caption{Example translation: Underlines show the matched constraints, and wavy lines show the phrases that the models fail to match.}\label{fig:example_translation}
\end{figure*}

Figure~\ref{fig:example_translation} shows the example translation of the baseline and LeCA with lexically constrained decoding.
Underlines in Figure~\ref{fig:example_translation} show the terms that match the constraints.
Obviously, the baseline model generated the same term repeatedly and failed to translate while all of the constraints were satisfied.
The baseline model appears to struggle with generating the constraint ``superconductivity single phase auto-transformer.''
One likely reason for this is that the baseline model generated a phrase that was quite similar to the constraint in the early phase (marked with a wavy line in Figure~\ref{fig:example_translation}), and thus the model considered the constraint as translated.

In contrast, LeCA+LCD successfully translated the sentence with the constraints.
We believe this is because the LeCA model correctly gives higher scores to the constraint phrases compared to the baselines, helping to generate a sentence with constraints.

\begin{figure}[t]
    \centering
    \includegraphics[width=\linewidth]{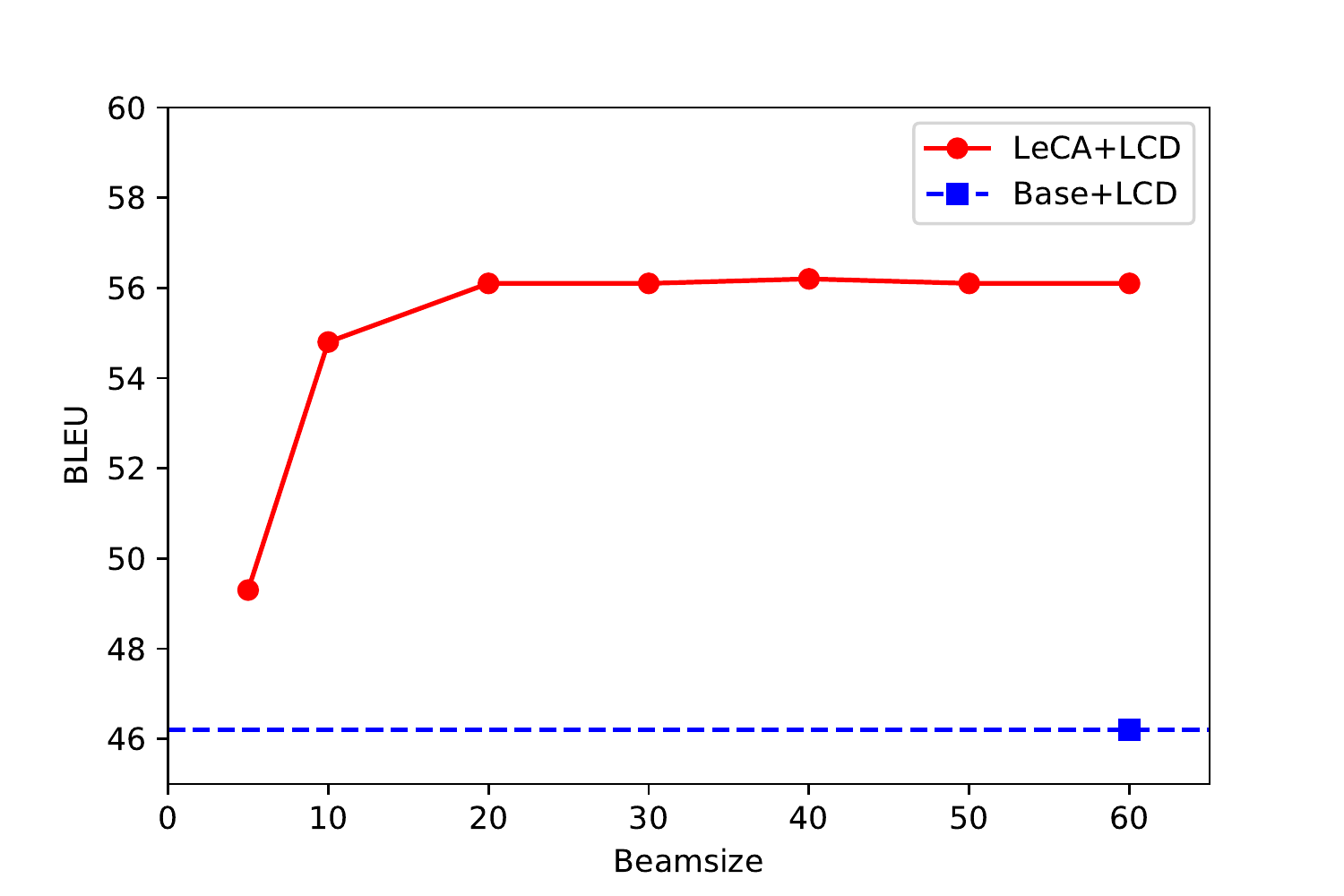}
    \caption{BLEU scores of \entoja\ translation decoding with various beam sizes. The BLEU scores are calculated with \texttt{sacreblue}~\cite{post18sacrebleu}}
    \label{fig:beamsize}
\end{figure}
Figure~\ref{fig:beamsize} shows the BLEU scores of \entoja\ translation decoding with various beam sizes.
As mentioned in \S\ref{sec:lcd}, the beam size of \baseline+LCD needs to be larger than 60 to successfully generate all of the constraints.
In contrast, LeCA+LCD can generate all of the constraints and improve the translation accuracy even when their beam size is quite small.
This result indicates that the output of LeCA is helpful for LCD to score the candidates and that LeCA can save inference time.

\section{Related Work}
\citet{hokamp-liu-2017-lexically} proposed Grid Beam Search (GBS), an extended beam search algorithm that forces the NMT model to output pre-specified lexical constraints of words or phrases.
At each decoding step, a beam is allocated to each number of constraints, and the top-k candidates that contain $n$ constraints are selected for the $n$\textsuperscript{th} beam.
Translations that satisfy the constraints appear in the beam corresponding to the number of constraints.
The beam size changes depending on the number of constraints for each sentence, which makes batch decoding difficult.
\citet{post-vilar-2018-fast} proposed Dynamic Beam Allocation (DBA), which dynamically allocates the beam with a fixed size and improves decoding more efficiently.
However, the distribution of the number of constraint tokens in the experiments of these papers was much smaller than that of this task, and we found these methods did not perform well on this task.

\citet{song2020alignment} and \citet{chen2021lexically} proposed lexically constrained decoding given explicit alignment guidance between the constraints and the source text.
Alignments were induced from an additional alignment head or attention weights \citep{garg-etal-2019-jointly}, but these methods assumed that gold alignments are given as constraints.
To apply these methods to this task, we would have to use an automatic alignment method (e.g., GIZA++, Fast-Align) to obtain the alignments, and the translation accuracy might suffer due to alignment error.

\citet{susanto-etal-2020-lexically} proposed non-autoregressive NMT for lexically constrained translation.
They used the Levenshtein Transformer \citep{gu-et-al-2019-levenshtein}, which inserts and deletes tokens at each time step, starting from the given constraints as the initial state.
They assumed that the order of the given constraints is the same as the order in the reference, but the given constraints in this task appear in random order.
Furthermore, they have not achieved comparable translation accuracy to the auto-regressive approaches.

Some works augment the input sequence with constraints.
\citet{song-etal-2019-code} augmented the source sentence by replacing or appending constraints with its corresponding source phrase through leveraging an SMT phrase table.
\citet{chen2020lexical_leca} proposed a simple yet effective augmentation method that appends constraints after the source sentence.
Although the decoding speed is fast, \citet{song-etal-2019-code} relied on the quality of the SMT phrase table.
Furthermore, neither of the works could guarantee that the translation would contains all constraints.

\section{Conclusion}
This paper described the systems that were submitted to the WAT 2021 restricted translation task.
We submitted systems for both \entoja\ and \jatoen, and both of our systems won the best translation accuracy as assessed by BLEU, the consistency score, and human evaluations.
We also confirmed that the data augmentation method makes lexically constrained decoding more effective and, furthermore, that combining data augmentation and constrained decoding significantly improves translation accuracy.

\bibliographystyle{acl_natbib}
\bibliography{myplain,main,acl2021}
\end{document}